\documentclass{article}
\usepackage{spconf,amsmath,epsfig}
\usepackage{multirow}
\usepackage{multicol}
\usepackage{booktabs}
\usepackage{gensymb}
\usepackage{makecell}
\usepackage{pgfplots}
\pgfplotsset{compat=newest}
\usepgfplotslibrary{groupplots}
\usepgfplotslibrary{dateplot}
\usepackage{booktabs}
\usepackage{tikzscale}
\usepackage{hyperref}
\usepackage{algorithmic}
\usepackage{array}
\usepackage{url}
\usepackage{amssymb}
\usepackage{graphicx}
\usepackage{tablefootnote}

\title{S$^3$Net: Innovating Stereo Matching and Semantic Segmentation with a Single-Branch Semantic Stereo Network in Satellite Epipolar Imagery}

\name{Qingyuan Yang, Guanzhou Chen\sthanks{Corresponding author: cgz@whu.edu.cn; zxdlmars@whu.edu.cn}, Xiaoliang Tan, Tong Wang, Jiaqi Wang, and Xiaodong Zhang\footnotemark[1]} 
\address{State Key Laboratory of Information Engineering in Surveying, Mapping and Remote Sensing,\\ 
Wuhan University, Wuhan 430079, China}

\begin{document}
%
\maketitle

\begin{abstract}

Stereo matching and semantic segmentation are significant tasks in binocular satellite 3D reconstruction. However, previous studies primarily view these as independent parallel tasks, lacking an integrated multitask learning framework. This work introduces a solution, the Single-branch Semantic Stereo Network (S$^3$Net), which innovatively combines semantic segmentation and stereo matching using Self-Fuse and Mutual-Fuse modules. Unlike preceding methods that utilize semantic or disparity information independently, our method identifies and leverages the intrinsic link between these two tasks, leading to a more accurate understanding of semantic information and disparity estimation. Comparative testing on the US3D dataset proves the effectiveness of our S$^3$Net. Our model improves the mIoU in semantic segmentation from 61.38 to 67.39, and reduces the D1-Error and average endpoint error (EPE)  in disparity estimation from 10.051 to 9.579 and 1.439 to 1.403 respectively, surpassing existing competitive methods. Our codes are available at: \href{https://github.com/CVEO/S3Net}{https://github.com/CVEO/S3Net}.

\end{abstract}


\section{Introduction}

Stereo matching, also known as disparity estimation, uses corrected epipolar images (binocular images) to determine depth information for 3D reconstruction and environmental perception. This is achieved by calculating the horizontal pixel offset of tie-points \cite{liao2022linear}. Various deep networks for image disparity estimation have achieved desirable results on RGB images, thanks to the rapid development of deep learning. However, these methods are susceptible to the data distribution of binocular images, which may result in training instability and confusion in disparity estimation. This limits their application in binocular or multi-view stereo satellite images \cite{bosch2019semantic}.


To address this issue, recent research has combined semantic segmentation and stereo matching tasks on satellite epipolar images. This has led to a new paradigm called satellite semantic stereo \cite{bosch2019semantic}. Semantic features of each pixel can effectively tackle issues such as blurred object disparity boundaries in disparity estimation. Meanwhile, disparity networks can help distinguish foreground and background, addressing a recurring challenge in semantic segmentation. Despite these advancements, most research treats stereo matching and semantic segmentation as separate tasks or focuses on improving their accuracy independently \cite{liao2023s}, leading to inadequate utilization of their close connection.

In this study, we introduce the end-to-end \textbf{S}ingle-branch \textbf{S}emantic \textbf{S}tereo Network (S$^3$Net), a novel approach that unifies semantic segmentation and disparity estimation to leverage the inherent correlation between semantic content and disparity. In doing so, it captures their inherent connection, thus improving semantic understanding and disparity accuracy. This closely coupled multi-task learning allows for a better understanding of complex scenes, consequently boosting robustness and generalizability.

\section{Methodology}

\begin{figure*}[htbp]
	\centering
  \includegraphics[width=0.95\linewidth]{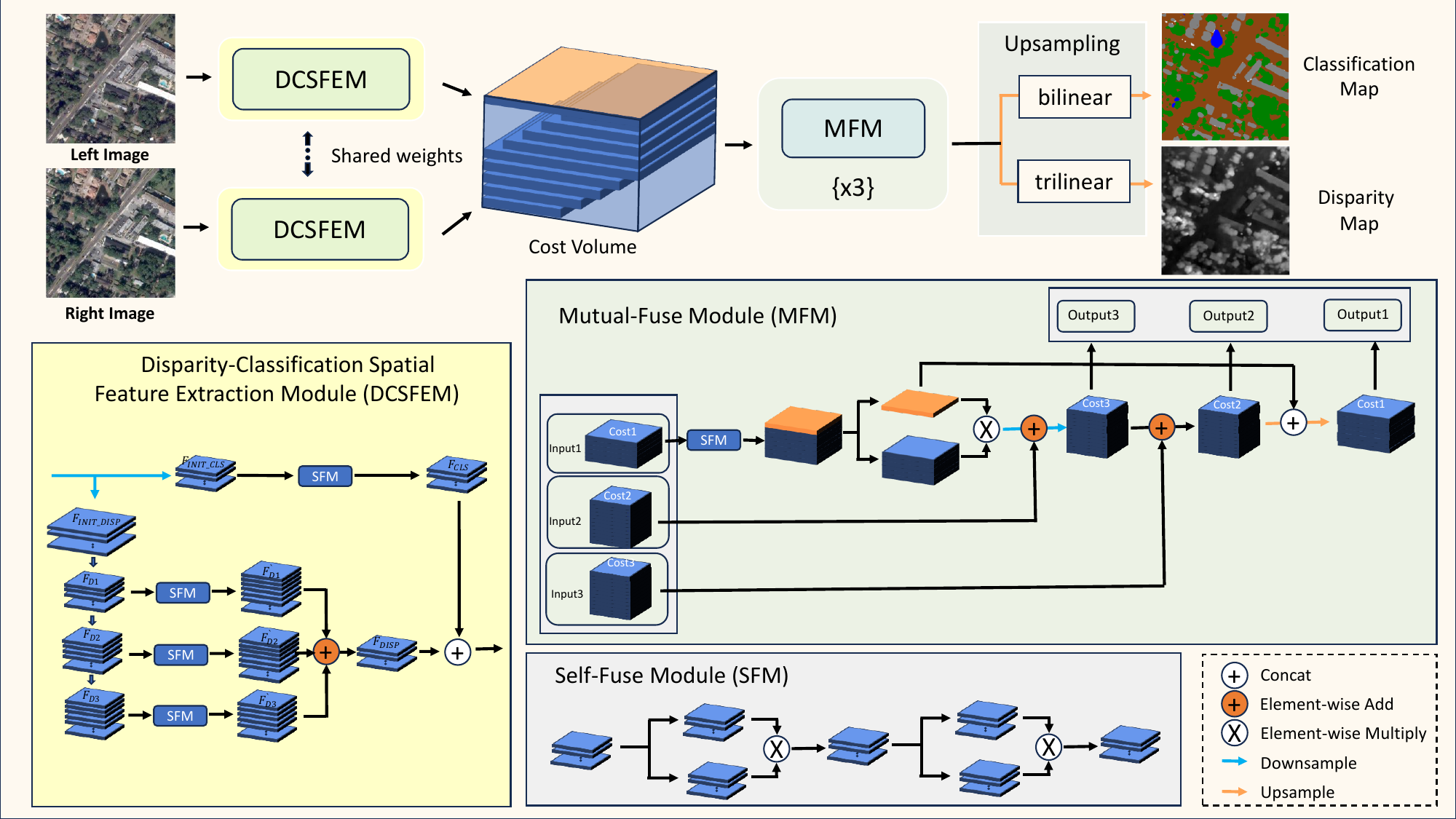}
  \caption{Framework of the Single-branch Semantic Stereo Network (S$^3$Net).}
  \label{fig_flow}
\end{figure*}

The overall architecture of S$^3$Net is shown in Fig.\ref{fig_flow}. Unlike traditional designs, our network uses a single branch configuration. It starts with the Disparity-Classification Spatial Feature Extraction Module (DCSFEM) that extracts features from the left and right images, generating a 4D cost volume containing semantic and disparity information. The Mutual-Fuse Module (MFM) then processes this volume, integrating disparity and semantic information. Finally, subjecting the cost volume to both trilinear and bilinear upsampling strategies results in two outputs at the original resolution: a disparity map and a pixel-level classification map.


\subsection{Disparity-Classification Spatial Feature Extraction Module (DCSFEM)}


We design a weight-sharing DCSFEM to merge semantic and disparity tasks, extracting features from both left and right images. This module consists of disparity and semantic feature extraction, using multi-scale and sequence processing strategies respectively. Both processes undergo four times downsampling. We introduce a Self-Fuse Module (SFM, see \ref{subsec:sfm}) for multi-scale disparity features, and concatenate the results with semantic features for synergy. The multi-scale features of the image pairs are then stacked to form a 4D cost volume.

\subsection{Cost Volume}

Unlike traditional cost volume stacking methods such as PSMNet \cite{chang2018pyramid} and S$^2$Net \cite{liao2023s}, we employ a selective approach towards stacking the multi-scale image features from both the left and right images, after they are processed through DCSFEM. The resultant structure forms a 4D cost volume (represented as $H \times W \times D \times C$) with dimensions corresponding to the height, width, number of disparities, and number of feature maps \label{subsec:feature}, which inherently includes an array of rich disparity and semantic features. The topmost layer of disparity in this 4D cost volume is reserved for semantic information, whereas the successive layers encapsulate disparity information from multi-scale features.

\subsection{Self-Fuse Module (SFM)}
\label{subsec:sfm}
To enhance the network's ability to handle noise interference in images and fully excavate intermediate layer information to more comprehensively characterize significant features in images, we have constructed an adaptive SFM module. This module can be divided into 2D and 3D types. Taking the 2D type as an example, it processes the input features through a dual-branch method, each branch applying a similar 2D convolution but with different weight parameters. The output features of the two branches are multiplied element-by-element according to their respective channels, and then output after the same operation. This module allows the network to adaptively control the information flow and achieve dynamic regulation and filtering on all feature information, thereby improving the network's expressive ability and learning efficiency, making the network more resistant to interference.

\subsection{Mutual-Fuse Module (MFM)}

This subsection details the use of 3D convolution operations in the MFM module to process three cost volumes (cost1, cost2, cost3) and output processed volumes. A total of three rounds need to be processed. In the first round, only cost1 (the initial cost volume) is inputted. The module begins with 3D SFM (mentioned in \ref{subsec:sfm}) processing on cost1, enabling the network to self-adjust information flow and capture practical information. This is followed by disparity dimension isolation, facilitating the fusion of semantic and disparity features. In order to better refine and integrate cost-volume information, we downsample the fused features and generate cost2 and cost3 at different stages via skip-connection, which serve as the input cost2 and cost3 for the next round. Finally, through upsampling, we restore the original shape and connect to the semantic layer as the input cost1 for the next round. After three rounds with different weights, we take the final cost1 as the input for subsequent processing.


\section{Experiments}

\subsection{Experimental settings}
We adopted the US3D dataset \cite{bosch2019semantic} for training and evaluation in this study. The dataset includes 4292 stereo image pairs of size $1024 \times 1024$, each with a classification and disparity map. We cropped 3500 images to $512 \times 512$ for training, used 338 for validation, and 454 for the test.
We adopted mIoU as the evaluation metric for semantic segmentation, EPE and D1-Error as the evaluation metrics for disparity estimation, and mIoU-3 \cite{bosch2019semantic} as the evaluation metric for considering both disparity and semantic segmentation performance.
We implemented our method based on the PyTorch 1.8.1 framework. When training these models, we set the batch size as four. All methods employed in this experiment were trained and tested on a workstation with Nvidia Tesla V100 16-GB GPUs.

\subsection{Ablation Study}

We evaluated our network's key modules (SFM, DCSFEM, and MFM) on the US3D dataset. We tested them separately, maintaining consistent dataset distribution. Our three-part evaluation included testing dual tasks without SFM, analyzing the disparity module in DCSFEM and MFM, and assessing the semantic module in DCSFEM and MFM. Table \ref{tab:ablation} shows improved dual task accuracy when SFM supports the integrated modules in DCSFEM and MFM.

\begin{table}[htbp]
  \centering
  \caption{Results of Ablation Study. (DM and SM represent the disparity module and semantic module in DCSFEM, DCV and SCV represent the disparity cost volume and semantic cost volume in MFM)}
    \resizebox{\linewidth}{!}{
    \begin{tabular}{ccccccccc}
    \toprule
    \multirow{2}[4]{*}{SFM} & \multicolumn{2}{c}{DCSFEM} & \multicolumn{2}{c}{MFM} & \multirow{2}[4]{*}{mIoU} & \multirow{2}[4]{*}{mIoU-3} & \multirow{2}[4]{*}{D1-Error} & \multirow{2}[4]{*}{EPE} \\
\cmidrule{2-5}          & DM    & SM    & DCV   & SCV   &       &       &       &  \\
    \midrule
    -     & $\checkmark$ & $\checkmark$ & $\checkmark$ & $\checkmark$ & 64.13 & 62.72 & 10.443 & 1.483 \\
    $\checkmark$ & $\checkmark$ & -     & $\checkmark$ & -     & -     & -     & 11.391 & 1.567 \\
    $\checkmark$ & -     & $\checkmark$ & -     & $\checkmark$ & 52.42 & -     & -     & - \\
    \midrule
    $\checkmark$ & $\checkmark$  & $\checkmark$ & $\checkmark$  & $\checkmark$ & 67.39 & 66.27  & 9.579     & 1.403 \\
    \bottomrule
    \end{tabular}}%
  \label{tab:ablation}%
\end{table}%

\subsection{Comparative Analysis with Other Methods}

\subsubsection{Compared Methods}

Our analysis primarily involves two different aspects: For the purpose of disparity estimation comparison, we conduct an exhaustive evaluation of currently superior algorithms such as PSMNet \cite{chang2018pyramid}, GwcNet \cite{guo2019group}, GANet \cite{zhang2019ga}, CFNet \cite{shen2021cfnet}, and S$^2$Net \cite{liao2023s}; In evaluating the task of semantic segmentation, we have selected advanced segmentation algorithms including SegFormer \cite{xie2021segformer}, PSPNet \cite{zhao2017pyramid}, SDFCNv2 \cite{chen2021sdfcnv2}, and HRNetV2 \cite{sun2019high}.

\subsubsection{Stereo Matching task}
\begin{figure*}[htbp]
	\centering
  \includegraphics[width=0.95\linewidth]{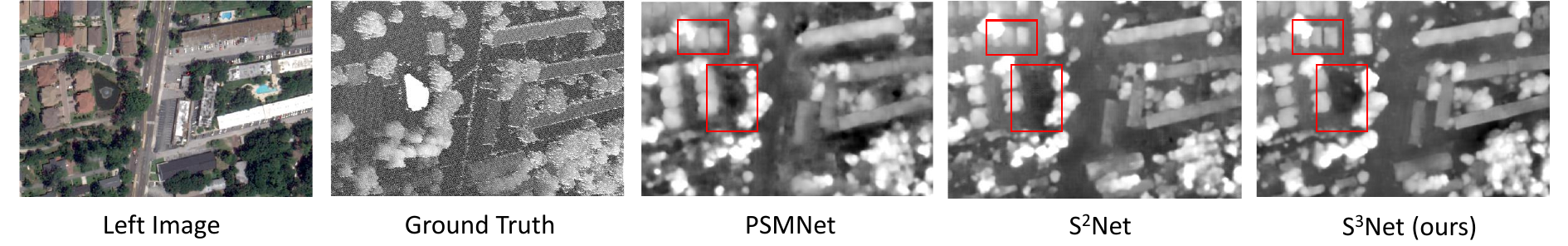}
  \caption{The comparison of S$^3$Net with other methods in disparity estimation tasks on the US3D dataset.}
  \label{fig_disp}
\end{figure*}

As shown in Table \ref{tab:stereo_matching}, our proposed S$^3$Net significantly outperformed other methods, demonstrating lower D1-Error (9.579) and EPE (1.403). As shown in Fig.\ref{fig_disp}, although PSMNet and S$^2$Net both showed good results, our method presented more detailed and accurate disparity details, especially at object edges and in areas with rich textures. As shown in the red box in the Fig.\ref{fig_disp}, our method better reflected the outline of the building and the edge information of the water. 

\begin{table}[htbp]
  \centering
  \caption{Results of stereo matching on the US3D test set}
  \resizebox{\linewidth}{!}{
    \begin{tabular}{ccccccc}
    \toprule
    Methods & PSMNet & GwcNet & GANet & CFNet & S$^2$Net & Ours \\
    \midrule
    D1-Error & 11.872 & 11.387 & 10.876 & 11.024 & 10.051 & \textbf{9.579} \\
    EPE   & 1.695 & 1.618 & 1.526 & 1.57  & 1.439 & \textbf{1.403} \\
    \bottomrule
    \end{tabular}}%
  \label{tab:stereo_matching}%
\end{table}%

\subsubsection{Semantic Segmentation task}
\begin{figure*}[htbp]
	\centering
  \includegraphics[width=0.95\linewidth]{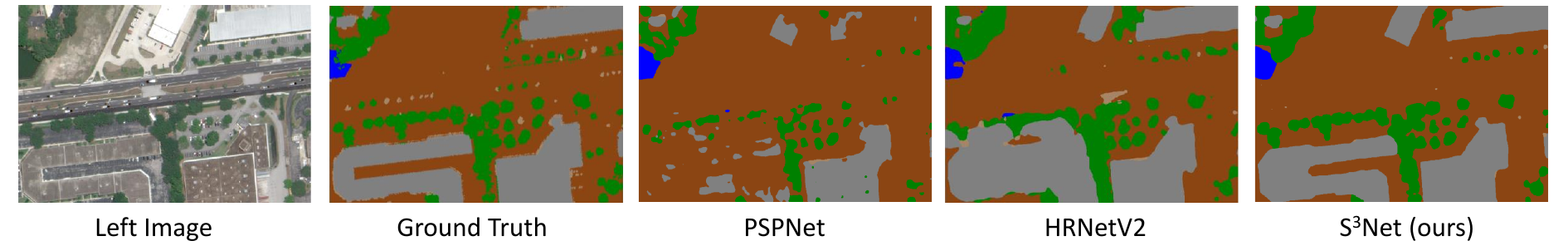}
  \caption{The comparison of S$^3$Net with other methods in semantic segmentation tasks on the US3D dataset.}
  \label{fig_cls}
\end{figure*}

According to the Table \ref{tab:semantic_segmentation}, our S$^3$Net demonstrates outstanding performance across different categories and performs better in specific scenarios (such as Water and Bridge). As shown in Fig.\ref{fig_cls}, although the PSPNet and HRNetV2 respectively show good results on water bodies and buildings, our method presents clearer contours for categories such as buildings, trees, and water.

\begin{table}[htbp]
  \centering
  \caption{Results of semantic segmentation on the US3D test set}
  \resizebox{\linewidth}{!}{
    \begin{tabular}{cccccc}
    \toprule
    Methods & SDFCNv2 & SegFormer & PSPNet & HRNetV2 & Ours \\
    \midrule
    Ground & 79.48  & 80.01  & 78.28  & 80.65  & \textbf{81.94 } \\
    Tree  & 64.88  & 64.47  & 59.32  & 65.53  & \textbf{66.39 } \\
    Building & 68.95  & 71.68  & 69.11  & 71.92  & \textbf{73.45 } \\
    Water & 65.28  & 59.44  & 68.82  & 68.27  & \textbf{79.23 } \\
    Bridge & 14.42  & 25.44  & 27.01  & 20.51  & \textbf{35.96 } \\
    \midrule
    mIoU  & 58.60  & 60.21  & 60.51  & 61.38  & \textbf{67.39 } \\
    \bottomrule
    \end{tabular}}%
  \label{tab:semantic_segmentation}%
\end{table}%



\section{Conclusion}
\label{sec_conclusion}

In the research, we introduce a novel multitask learning framework called the (S$^3$Net) to simultaneously infer disparity maps and classification maps. The uniqueness of our method stems from capitalizing on the strong correlation between these tasks, effectively integrating them via self-fusion and mutual fusion modules for mutual enhancement. Notably, the evaluation results obtained from the US3D dataset and the comparison with other models affirm the feasibility and exceptional performance of our task framework. In the future, we hope to extend the results of this study to applications in multiview stereo matching and 3D reconstruction of multi-sensor data, and further expand the experimentation of this method in various imagery scenarios.

\section{Acknowledgement}
\label{sec_ack}

This research was funded by the National Natural Science Foundation of China (No.42101346), the China Postdoctoral Science Foundation (No.2020M680109), and the Wuhan East Lake High-tech Development Zone Program of Unveiling and Commanding (No.2023KJB212). 




\bibliographystyle{IEEEbib}
\bibliography{refs}

\begin{thebibliography}{10}

\bibitem{liao2022linear}
Puyun Liao, Guanzhou Chen, Xiaodong Zhang, Kun Zhu, Yuanfu Gong, Tong Wang, Xianwei Li, and Haobo Yang,
\newblock ``A linear pushbroom satellite image epipolar resampling method for digital surface model generation,''
\newblock {\em ISPRS Journal of Photogrammetry and Remote Sensing}, vol. 190, pp. 56--68, 2022.

\bibitem{bosch2019semantic}
Marc Bosch, Kevin Foster, Gordon Christie, Sean Wang, Gregory~D Hager, and Myron Brown,
\newblock ``Semantic stereo for incidental satellite images,''
\newblock in {\em 2019 IEEE Winter Conference on Applications of Computer Vision (WACV)}. IEEE, 2019, pp. 1524--1532.

\bibitem{liao2023s}
Puyun Liao, Xiaodong Zhang, Guanzhou Chen, Tong Wang, Xianwei Li, Haobo Yang, Wenlin Zhou, Chanjuan He, and Qing Wang,
\newblock ``S2net: A multitask learning network for semantic stereo of satellite image pairs,''
\newblock {\em IEEE Transactions on Geoscience and Remote Sensing}, vol. 62, pp. 1--13, 2024.

\bibitem{chang2018pyramid}
Jia-Ren Chang and Yong-Sheng Chen,
\newblock ``Pyramid stereo matching network,''
\newblock in {\em Proceedings of the IEEE conference on computer vision and pattern recognition}, 2018, pp. 5410--5418.

\bibitem{guo2019group}
Xiaoyang Guo, Kai Yang, Wukui Yang, Xiaogang Wang, and Hongsheng Li,
\newblock ``Group-wise correlation stereo network,''
\newblock in {\em Proceedings of the IEEE/CVF conference on computer vision and pattern recognition}, 2019, pp. 3273--3282.

\bibitem{zhang2019ga}
Feihu Zhang, Victor Prisacariu, Ruigang Yang, and Philip~HS Torr,
\newblock ``Ga-net: Guided aggregation net for end-to-end stereo matching,''
\newblock in {\em Proceedings of the IEEE/CVF Conference on Computer Vision and Pattern Recognition}, 2019, pp. 185--194.

\bibitem{shen2021cfnet}
Zhelun Shen, Yuchao Dai, and Zhibo Rao,
\newblock ``Cfnet: Cascade and fused cost volume for robust stereo matching,''
\newblock in {\em Proceedings of the IEEE/CVF Conference on Computer Vision and Pattern Recognition}, 2021, pp. 13906--13915.

\bibitem{xie2021segformer}
Enze Xie, Wenhai Wang, Zhiding Yu, Anima Anandkumar, Jose~M Alvarez, and Ping Luo,
\newblock ``Segformer: Simple and efficient design for semantic segmentation with transformers,''
\newblock {\em Advances in Neural Information Processing Systems}, vol. 34, pp. 12077--12090, 2021.

\bibitem{zhao2017pyramid}
Hengshuang Zhao, Jianping Shi, Xiaojuan Qi, Xiaogang Wang, and Jiaya Jia,
\newblock ``Pyramid scene parsing network,''
\newblock in {\em Proceedings of the IEEE conference on computer vision and pattern recognition}, 2017, pp. 2881--2890.

\bibitem{chen2021sdfcnv2}
Guanzhou Chen, Xiaoliang Tan, Beibei Guo, Kun Zhu, Puyun Liao, Tong Wang, Qing Wang, and Xiaodong Zhang,
\newblock ``Sdfcnv2: An improved fcn framework for remote sensing images semantic segmentation,''
\newblock {\em Remote Sensing}, vol. 13, no. 23, pp. 4902, 2021.

\bibitem{sun2019high}
Ke~Sun, Yang Zhao, Borui Jiang, Tianheng Cheng, Bin Xiao, Dong Liu, Yadong Mu, Xinggang Wang, Wenyu Liu, and Jingdong Wang,
\newblock ``High-resolution representations for labeling pixels and regions,''
\newblock {\em arXiv preprint arXiv:1904.04514}, 2019.

\end{thebibliography}

\end{document}